\documentclass[letterpaper, 10 pt, conference]{ieeeconf}

\IEEEoverridecommandlockouts
\usepackage{amsmath}
\usepackage{amssymb}
\usepackage{booktabs}
\usepackage{cite}
\usepackage{graphicx}
\usepackage{multirow}
\usepackage{capt-of}
\usepackage{url}
\usepackage{xcolor}
\usepackage{tabularx}
\newcolumntype{Y}{>{\centering\arraybackslash}X}
\newcolumntype{L}{>{\raggedright\arraybackslash}X}

\makeatletter
\renewcommand{\fnum@table}{TABLE~\thetable}
\long\def\@makecaption#1#2{%
\ifx\@captype\@IEEEtablestring%
\par\noindent{\footnotesize #1.~~\normalfont #2}\par%
\@IEEEtablecaptionsepspace%
\else%
\@IEEEfigurecaptionsepspace%
\setbox\@tempboxa\hbox{\footnotesize #1.~~ #2}%
\ifdim \wd\@tempboxa >\hsize%
\setbox\@tempboxa\hbox{\footnotesize #1.~~ }%
\parbox[t]{\hsize}{\footnotesize \noindent\unhbox\@tempboxa#2}%
\else%
\ifcenterfigcaptions \hbox to\hsize{\footnotesize\hfil\box\@tempboxa\hfil}%
\else \hbox to\hsize{\footnotesize\box\@tempboxa\hfil}%
\fi\fi\fi}
\makeatother

\title{\LARGE \bf
SARI: Phase-Split Sim-Real Co-Training for Contact-Rich Manipulation
}
\author{Xingxin He$^{1}$, Yuxuan Jiang$^{1}$, Haonan Zhang$^{1}$, Chuhan Cui$^{1}$,\\
Kaile Li$^{1}$, Zhongxing Zheng$^{2}$, Caihao Xu$^{2}$, Ziqi Wang$^{1,*}$\\[0.2em]
$^{1}$The Hong Kong University of Science and Technology\quad
$^{2}$BYD Company Limited
}

\IEEEaftertitletext{%
\begin{minipage}{\textwidth}
    \centering
    \includegraphics[width=1.0\linewidth]{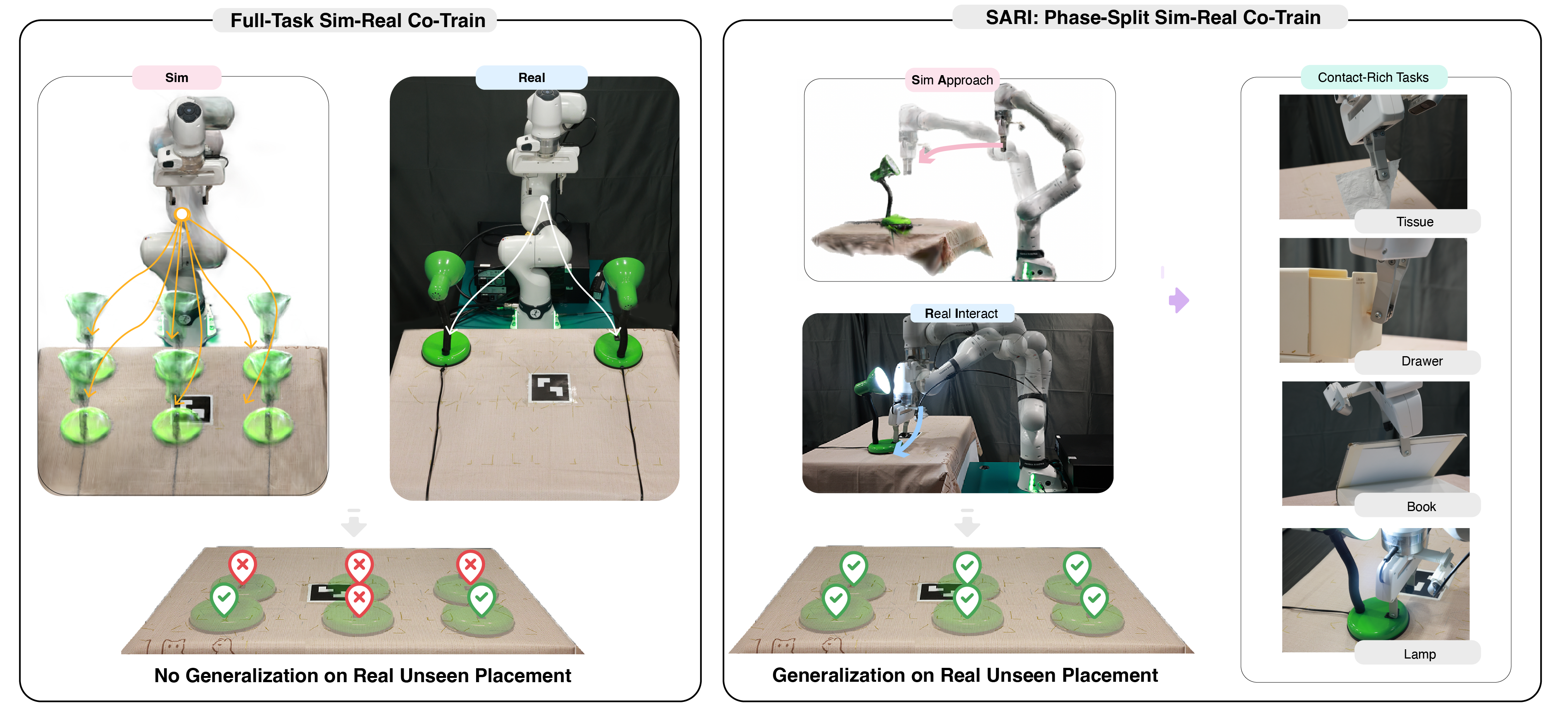}
    \captionof{figure}{
    \textbf{SARI (Simulated Approach, Real Interaction)} allocates each phase of a
    contact-rich task to the domain best suited to provide its supervision. Unlike
    full-task sim-real co-training (left), which requires end-to-end data in both
    domains and can fail at unseen real placements, SARI (right) combines
    workspace-diverse simulated approaches with real contact interactions. The
    resulting phase-split dataset post-trains a single policy to complete tasks at
    unseen placements across deformable, frictional, articulated, and spring-loaded
    interactions. Our project website: https://icra27-sari.vercel.app }
    \label{fig:teaser}
\end{minipage}
\par\vspace{0.25\baselineskip}
}

\begin{document}

\maketitle
\thispagestyle{empty}
\pagestyle{empty}

\begin{abstract}
Vision-language-action (VLA) models often require costly real-world demonstrations to adapt to contact-rich manipulation tasks, particularly when generalization across object placements is needed. We propose \textbf{SARI} (Simulated Approach, Real Interaction), a phase-split sim-and-real co-training framework built on a simple insight: spatial coverage and contact physics should be acquired from the domains best suited to them. Specifically, free-space approaches require spatial diversity but tolerate modest simulation gaps, making them ideal for synthetic generation; conversely, contact interactions demand accurate physics but vary little across object placements, allowing a few real demonstrations to generalize across the workspace. SARI generates diverse simulated approaches in a photorealistic digital twin while collecting real contact interactions at only a few placements. Post-trained on these phase-segmented demonstrations, a single policy seamlessly stitches simulated approaches with real contact interactions using visual appearance alignment and a shared camera-relative action representation—without explicit phase labels or hand-coded switches. Across five real-world contact-rich manipulation tasks, SARI reduces real-data collection time by 34.3\% and achieves 27.5\% success at unseen placements, where all full-task sim-real baselines fail completely (0\%).
\end{abstract}

\section{INTRODUCTION}
\label{sec:introduction}

Vision-language-action (VLA) models have demonstrated strong and 
increasingly general manipulation capabilities, but this progress 
has been enabled by human-collected robot data at a formidable scale. 
Recent VLAs, including $\pi_{0.5}$~\cite{pi05}, GR00T N1.7~\cite{gr00tn1}, 
and GEN-1.5~\cite{gen15}, are trained on large and diverse 
collections of robot, human, and synthetic data. Yet adapting these models 
to a new, contact-rich task still commonly requires many task demonstrations, 
and long-horizon tasks amplify this demand because errors and distribution 
shifts accumulate over time.

An important cause is the limited spatial generalization of VLA policies. 
Although visual encoders are pretrained on broad data, post-trained policies 
can still overfit an action sequence to the specific object placements observed in 
their demonstrations, rather than reliably extrapolating across the 
workspace~\cite{demogen,r2rgen}. Consequently, achieving workspace-wide coverage requires the redundant repetition of near-identical free-space approaches, rather than focusing effort on contact-rich supervision.

Simulation offers an effective route to alleviate this human collection burden. 
Recent advances, including MimicGen~\cite{mimicgen},
DemoGen~\cite{demogen}, Real2Render2Real~\cite{real2render2real}, and
others~\cite{robosplat,simandreal}, synthesize additional demonstrations by transforming a small set of
source demonstrations, generating trajectories in reconstructed scenes or combining simulated
and real data for policy training.
These methods can efficiently expand workspace coverage, diversify trajectories, and 
eliminate redundant teleoperation of identical free-space approaches.

However, existing synthetic-data pipelines commonly assume that simulation is reliable throughout an entire task or require accurate modeling of contact 
dynamics~\cite{dexmimicgen,demogen,robosplat,simandreal}. 
In contact-rich manipulation, unmodeled friction, compliance, object resistance, 
and visual sim-to-real discrepancies can make synthetic contact data unreliable. 
To bridge this gap, straightforward sim-and-real co-training mixes full-task simulated and real demonstrations; yet, this still wastes costly real-world effort on free-space approaches and allows the spatially narrow real-data distribution to override and dilute the broad spatial coverage provided by simulation.

To address these challenges, we propose \textbf{SARI} (Simulated Approach, Real Interaction), a phase-split co-training framework built on a simple insight: spatial coverage and contact physics should be acquired from the domains best suited to them. 
Simulation can cheaply provide broad spatial coverage for free-space approaches, while a few real demonstrations suffice to teach contact interactions that generalize across placements.
Specifically, the approach phase requires broad spatial diversity but tolerates modest simulation gaps, making it ideal for synthetic generation. Conversely, the interaction phase demands accurate contact dynamics but varies little across object placements, allowing a modest number of real demonstrations at a few placements to generalize across the workspace.
Figure~\ref{fig:teaser} contrasts this phase-split allocation with full-task sim-real co-training.

We instantiate this phase-specific allocation using a Gaussian-Splatting~\cite{gaussiansplatting} 
digital twin that generates simulated approaches across all placements, while collecting real contact interactions near the pre-contact boundary at only a subset. To enable a single policy to learn seamlessly from both sources without phase labels or a hand-coded switch, we align visual appearances and express actions in a shared camera-relative representation. 
Across five real-world contact-rich manipulation tasks, \textbf{SARI} reduces real-data collection time by 34.3\% and achieves 27.5\% unseen task completion, whereas all baselines achieve 0\%.

Our core contributions are:
\begin{itemize}
    \item We introduce \textbf{SARI}, a phase-split sim-and-real co-training
      framework that decouples workspace-wide spatial reaching from contact interactions, allocating data collection to the domain best suited for each phase.
    \item We present a sim-to-real alignment pipeline, enabling a single policy to seamlessly stitch simulated approaches with real contact interactions without explicit phase labels or switching heuristics.
    \item We evaluate \textbf{SARI} on a Franka robot across five contact-rich manipulation tasks, demonstrating that our camera-relative action representation serves as the key mechanism to unlock zero-shot spatial generalization to unseen placements while reducing real-world collection time.
\end{itemize}

\section{RELATED WORKS}
\label{sec:related-work}

\textbf{Synthetic data generation.}
Reconstructed scenes and digital twins are increasingly used to reduce
robot-data collection cost. Gaussian Splatting enables photorealistic
robot-workspace rendering and has recently supported simulation, memory,
and reactive control~\cite{gaussiansplatting,splatctrl,gaussmemory};
related systems synthesize demonstrations from few real examples or
reconstructed workspaces~\cite{mimicgen,dexmimicgen,demogen,real2render2real,robosplat,robosimgs,gsworld}.
However, these pipelines often require full-task demonstrations
or assume that synthetic rollouts are reliable throughout the task,
including contact regimes where friction, compliance, and object
resistance are hard to model. We therefore restrict reconstruction-based synthesis to the phase where geometric fidelity is sufficient: generating appearance-aligned free-space approaches across placements, while reserving contact-rich interaction for real data.

\textbf{Sim-and-real co-training.}
Sim-to-real and mixed-domain training combine simulation scalability with real-world deployment fidelity. Existing methods typically co-train policies on full-task simulated and real data~\cite{simandreal,cotrainmechanistic}, introduce online corrections or real-world residuals for contact-rich transfer~\cite{transic,sparr}, or study robust assembly in simulation prior to deployment~\cite{industreal,cips}. However, these approaches generally assume full-task trajectory in both domains or require extra adaptation machinery when simulated physics fails. In contrast, \textbf{SARI} exploits a fundamental task-phase asymmetry: simulation provides workspace-wide coverage for free-space approaches, while real demonstrations supply accurate contact dynamics at a sparse subset of placements. A single policy is post-trained on their union without phase labels, residual controllers, or phase-specific losses.

\textbf{VLA post-training.}
Large robot datasets and VLA policies have
made post-training a practical mechanism for adapting pretrained robot
models to new tasks~\cite{rt1,rt2,openxembodiment,octo,openvla,pi0,pi05,droid}.
Recent work improves VLA action generation, pretraining sources,
obstacle-aware imitation, and world-model conditioning~\cite{imlevla,huro,obstadiff,wmcraftnet},
while contact-rich variants introduce force-aware modules or
curricula~\cite{forcevla,craft}. Yet adapting VLAs to a new workspace
usually still requires full task demonstrations covering both approach
and interaction, and specialized contact signals add assumptions beyond
standard behavior cloning. We keep the recipe plain: a pretrained VLA, a shared camera-relative action representation, and phase-specific sim-real data. Together, these ingredients enable a single policy to seamlessly stitch simulated approach trajectories with real contact interactions, supporting spatial generalization to placements without real interaction demonstrations using only standard behavior cloning.

\section{PRELIMINARIES}
\label{sec:problem}

\begin{figure*}[t]
    \centering
    \includegraphics[width=\textwidth]{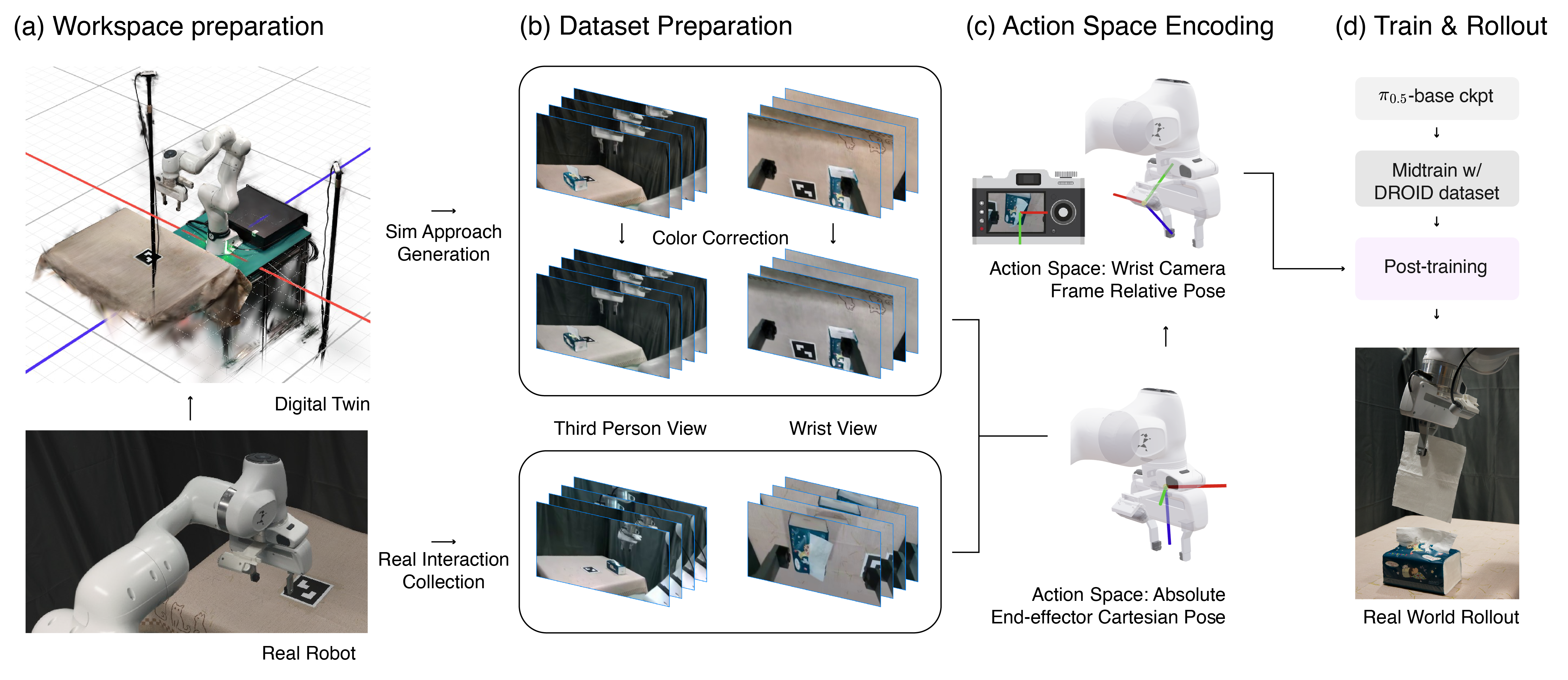}
    \caption{Overview of \textbf{SARI}. We reconstruct a digital twin of the real workspace,
    generate spatially diverse simulated approach segments, and collect real contact
    segments. Appearance-aligned observations and a shared camera-relative action
    representation allow a single policy to post-train on their union and execute the
    full task in the real world.}
    \label{fig:pipeline}
\end{figure*}

\subsection{Task setting and pre-contact boundary}
We consider single-object manipulation tasks whose demonstrations have
two segments. In the \emph{free-space segment}, the robot moves from
its deployment home toward the object without making contact. In the
\emph{contact segment}, it presses, pulls, or pushes the object to
complete the task. The segments meet at a \emph{pre-contact pose}: a
pose near the object that serves as the end of the simulated approach
and the starting region for real contact collection. Figure~\ref{fig:pipeline}
illustrates this setup: the robot travels from home, through free space, to
this pre-contact pose, and then executes the contact interaction.

To transfer a single calibrated pre-contact pose across the candidate object
placements, we anchor it to five coordinate frames spanning the robot, table,
object, and cameras. We write ${}^A X^B \in SE(3)$ for the pose of frame $B$ measured in
frame $A$, with rotation ${}^A R^B$ and translation ${}^A p^B$.
The task uses the
robot-base frame $W$, the table frame $T$, the object's canonical body
frame $O$, the physical end-effector flange frame $E$, and the wrist
camera optical frame $C$. The camera is
rigidly mounted to the end effector with calibrated transform
${}^E X^C$, so ${}^W X^C={}^W X^E\,{}^E X^C$.

Let $\mathcal{P}=\{1,\ldots,N\}$ index the candidate object placements.
The table has a fixed pose ${}^W X^T$, and each nominal placement
${}^T X^O_{(i)}$ keeps the object at its canonical yaw. We select one
object-relative pre-contact transform ${}^O X^E_{\mathrm{pre}}$ at a
reference placement and transfer it to the others:
\begin{equation}
{}^W X^E_{(i)}
= {}^W X^T\,{}^T X^O_{(i)}\,{}^O X^E_{\mathrm{pre}},
\qquad i\in\mathcal{P}.
\label{eq:prepress}
\end{equation}
We assume that the same object-relative pre-contact pose is suitable
at each placement. With the fixed wrist camera mounting, this also
preserves the nominal object pose in the optical frame:
\begin{equation}
{}^C X^O_{\mathrm{pre}}
= \left({}^O X^E_{\mathrm{pre}}\,{}^E X^C\right)^{-1}.
\label{eq:precontact-optical}
\end{equation}
We compare our phase-split the end-effector--object
relationship, the wrist camera encounters similar local views of the
object just before contact across placements. This geometric consistency
does not imply identical backgrounds, lighting, or contact dynamics.
Equation~\eqref{eq:prepress} specifies an approach target; it does not
specify the forces or motion needed to complete the contact behavior.

\subsection{Segmented, partially real dataset}
Let $\mathcal{P}_{\mathrm{real}}\subset\mathcal{P}$ be the subset of
placements where real contact demonstrations are collected. The
training set contains two sources:
\begin{itemize}
    \item $\mathcal{D}_{\mathrm{sim}}$: simulated free-space episodes
    from the home region to ${}^W X^E_{(i)}$, for every
    $i\in\mathcal{P}$;
    \item $\mathcal{D}_{\mathrm{real}}$: real contact episodes starting
    near ${}^W X^E_{(i)}$ and completing the interaction, only for
    $i\in\mathcal{P}_{\mathrm{real}}$.
\end{itemize}
Both sources use the same language instruction for a given task and the
same action convention. Neither source contains a home-to-contact full-task demonstration. We post-train one policy $\pi_\theta$
on their union, without supplying a phase label or an intermediate-goal
instruction.

\subsection{Evaluation regimes}
At evaluation, the robot starts from home and completes the task in a single continuous real-world rollout. We categorize object placements based on real-world demonstration coverage: \textbf{seen} placements ($\mathcal{P}_{\mathrm{real}}$) contain both simulated approaches and real contact demonstrations, while \textbf{unseen} placements ($\mathcal{P}\setminus\mathcal{P}_{\mathrm{real}}$) are covered strictly by simulated approaches. The key evaluation question is whether contact behaviors can generalize to unseen placements without any real interaction demonstrations.

\section{METHODS}
\label{sec:method}

Contact-rich manipulation requires wide spatial coverage and
accurate interaction dynamics, but real demonstrations collection
is costly and simulated contact unreliable. To address this mismatch, we propose \textbf{SARI}, a phase-split sim-real co-training method. As shown
in Figure~\ref{fig:pipeline}, \textbf{SARI} decouples the task execution into two phases: offloading free-space approaches to a workspace digital twin and collecting real contact segments from transferred pre-contact poses. By establishing observation alignment and a shared camera-relative action space, SARI fuses the simulated approach and real contact data to train a unified policy via imitation learning.

\subsection{Cross-domain appearance alignment}
\label{sec:appearance-alignment}
We build on GSWorld~\cite{gsworld} to render the reconstructed workspace
with 3D Gaussian Splatting~\cite{gaussiansplatting}. A motion planner
provides the robot trajectory, and the renderer provides observations
from camera poses calibrated to the real setup. The reconstruction must
cover the views encountered from home through the pre-contact region;
missing geometry can otherwise leave visible holes in the training
images.

Two changes improve the visual match:
\begin{itemize}
    \item \textbf{Higher-order spherical harmonics.} We improve the
    rendering component to preserve and evaluate degree-3 spherical-harmonics
    coefficients ($L=3$). This retains view-dependent appearance that
is lost when the higher-order terms are omitted.
    \item \textbf{Photometric color calibration.} We fit a separate
    color mapping \cite{real2simeval} for each camera using color statistics from selected
    material regions in simulated and real reference images. The
    references need not be paired frames of the same motion. We apply
    the fitted maps to simulated images while preserving their actions,
    robot states, and episode boundaries.
\end{itemize}
The observations include two exterior views and one wrist view. Color
calibration reduces appearance differences but cannot repair missing
geometry or incorrect camera poses. Section~\ref{sec:exp-ablation}
measures the contribution of each correction and tests sensitivity to a
synthetically corrupted GS scene.

Our exploration with NVIDIA Cosmos~\cite{cosmos} showed that generative
video translation can fill visual holes in rendered observations and
improve sim-real appearance alignment. However, applying it across a
demonstration dataset is computationally expensive. Rebuilding
the Gaussian scene to improve reconstruction quality is more efficient.

\subsection{Camera-relative action representation}
\label{sec:camera-actions}
Both simulated and real demonstrations record absolute Cartesian end-effector
commands, and we convert to camera-relative actions for policy
training. Intuitively, each action is re-expressed as a translation and
rotation relative to the robot's wrist camera view rather than
in absolute world coordinates, so that simulated and real trajectories
share one viewpoint-centered action space.

Using the frames defined in Section~\ref{sec:problem}, let ${}^W X^{E_k}$ be the
measured end-effector pose at row $k$ and ${}^W X^{U_j}$ the desired
end-effector pose at row $j$, both in the robot-base frame $W$.
Training labels are constructed from a sampled chunk of $H$ commands
$U_k,\ldots,U_{k+H-1}$, each expressed relative to the measured
end-effector pose and wrist-camera orientation at the start of the
chunk. Writing $p_k={}^W p^{E_k}$, $R_k={}^W R^{E_k}$,
$p^U_j={}^W p^{U_j}$, $R^U_j={}^W R^{U_j}$, and camera orientation
$R_{C,k}={}^W R^{C_k}=R_k\,{}^E R^C$, we follow the camera-frame
relative-motion formulation of E2VLA~\cite{e2vla} and define, for
$i=0,\ldots,H-1$,
\begin{equation}
\begin{aligned}
\delta p_k^{(i)} &= R_{C,k}^{\mathsf T}(p^U_{k+i}-p_k),\\
\delta R_k^{(i)} &= R_{C,k}^{\mathsf T}R^U_{k+i}R_k^{\mathsf T}R_{C,k}.
\end{aligned}
\label{eq:camera-forward}
\end{equation}
All targets in the chunk are expressed relative to this same starting
pose and camera orientation, rather than as successive step-to-step
increments.

We encode each training action as
\begin{equation}
a_k^{(i)}=
\bigl[(\delta p_k^{(i)})^{\mathsf T},\,
\rho_6(\delta R_k^{(i)})^{\mathsf T},\,g^U_{k+i}\bigr]^{\mathsf T}
\in\mathbb{R}^{10},
\label{eq:camera-action}
\end{equation}
where $\rho_6$ stacks the first two rotation-matrix rows in row-major
order~\cite{rot6d}, and $g^U\in[0,1]$ is absolute gripper closedness
($0$ open, $1$ closed). These normalized camera-relative labels give
both data sources the same action semantics in the wrist-view axes.

At rollout, we recover absolute Cartesian targets for the controller by
algebraically inverting Eq.~\eqref{eq:camera-forward} after undoing
normalization and decoding the predicted Rot6D rotations via
Gram--Schmidt orthonormalization; the gripper command passes through
unchanged. The anchor pose $p_k,R_k$ and camera orientation $R_{C,k}$
are read once at policy invocation and held fixed while decoding and
executing that chunk, then refreshed from the latest measurement at
the next invocation.

We midtrain a $\pi_{0.5}$ checkpoint~\cite{pi05} on
DROID~\cite{droid} converted to this action representation for 100,000
steps, then post-train it on the segmented task data so both use the
same action semantics. Section~\ref{sec:exp-action} compares this
checkpoint-and-representation combination with $\pi_{0.5}$-DROID joint
velocity, $\pi_{0.5}$-DROID joint position, and GR00T N1.7-DROID
end-effector-relative Cartesian actions, all post-trained on our
segmented task data.

\begin{table*}[!t]
\centering
\small
\setlength{\tabcolsep}{5pt}
\renewcommand{\arraystretch}{1.15}

\caption{\textbf{Lamp-task performance (\%).}
Each object placement is evaluated in 10 trials.
\textit{Approach} denotes approaching the pre-contact region, and
\textit{Completion} denotes successful completion of the lamp-pressing task.}

\label{tab:main-results}

\begin{tabular*}{\textwidth}
{@{\extracolsep{\fill}}llccccccrr@{}}
\toprule
Method & Metric
& \multicolumn{2}{c}{Seen}
& \multicolumn{4}{c}{Unseen}
& \multicolumn{2}{c}{Mean} \\
\cmidrule(lr){3-4}
\cmidrule(lr){5-8}
\cmidrule(lr){9-10}

& & P1 & P3 & P2 & P4 & P5 & P6 & Seen & Unseen \\
\midrule

\multirow{2}{*}{Full Sim}
& Approach   & 100 & 100 & 100 & 100 & 100 & 100 & 100.0 & 100.0 \\
& Completion & 0   & 0   & 0   & 0   & 0   & 0   & 0.0   & 0.0 \\
\addlinespace

\multirow{2}{*}{Full Real}
& Approach   & 100 & 100 & 100 & 0   & 0   & 0   & 100.0 & 25.0 \\
& Completion & 50  & 50  & 0   & 0   & 0   & 0   & 50.0  & 0.0 \\
\addlinespace

\multirow{2}{*}{Full-task Co-training~\cite{simandreal}}
& Approach   & 100 & 100 & 100 & 0   & 0   & 0   & 100.0 & 25.0 \\
& Completion & 50  & 60  & 0   & 0   & 0   & 0   & 55.0  & 0.0 \\
\addlinespace

\multirow{2}{*}{\textbf{SARI (Ours)}}
& Approach   & 100 & 100 & 100 & 100 & 100 & 100 & 100.0 & 100.0 \\
& Completion & 50  & 50  & 40  & 20  & 30  & 20  & 50.0  & \textbf{27.5} \\

\bottomrule
\end{tabular*}
\end{table*}

\begin{table}[t]
    \centering
    \caption{\textbf{Data allocation and real collection cost for the lamp task.}
    \textbf{SARI} uses simulated approaches and real contacts; all methods use matched
    initialization, post-training compute, and evaluation. 
    }
    \label{tab:episode-composition}

    \begin{tabularx}{\columnwidth}{
    @{}
    >{\raggedright\arraybackslash}p{0.23\columnwidth}
    >{\centering\arraybackslash}X
    cc
    @{}}
        \toprule
        Method & Sim. demos & Real demos & Real time (min) \\
        \midrule
        Full Sim & \mbox{180 full-task} & -- & 0 \\
        Full Real & -- & \mbox{60 full-task} & 28.03 \\
        \shortstack[l]{Full-task\\Co-training~\cite{simandreal}}
        & \mbox{180 full-task} & \mbox{60 full-task} & 28.03 \\
        \textbf{SARI (Ours)}
        & \mbox{\textbf{180 approaches}}
        & \mbox{\textbf{60 contacts}}
        & \textbf{18.42} \\
        \bottomrule
    \end{tabularx}
\end{table}

\subsection{Segmented data curation}
\label{sec:segmented-generation}
For each task, we first define ${}^O X^E_{\mathrm{pre}}$ at one
reference placement. We transfer it to all candidate placements using
Eq.~\eqref{eq:prepress} and generate collision-free approaches from the
deployment home region. These episodes stop when the robot reaches the
pre-contact region; they contain no simulated contact and no appended
terminal hold. Their images pass through the appearance-alignment
procedure above.

For real-world data $\mathcal{P}_{\mathrm{real}}$ collection, the robot is programmatically initialized at the transferred pre-contact pose ${}^W X^E_{(i)}$, which serves as the starting pose. The operator then completes the remaining contact interaction via teleoperation. The same construction applies to all tasks shown in Figure~\ref{fig:tasks} with different object assets and pre-contact transforms.

\subsection{Training}
We post-train one $\pi_{0.5}$ policy on
$\mathcal{D}_{\mathrm{sim}}\cup\mathcal{D}_{\mathrm{real}}$ with
standard flow-matching behavior cloning. The policy receives camera
observations, robot state, and one language instruction for the task.
It predicts action chunks and is deployed from home without a
hand-coded phase switch, phase-dependent loss weights, or an auxiliary
alignment loss.

The sim-real sampling ratio controls how often each source contributes
training samples. It is a training setting, separate from the number
of recorded episodes; we give both mixed-data methods the same
validation budget for choosing it. No additional weighting is based on
the task phase or distance to the pre-contact pose. The post-training
schedule, sampling ratio, and prediction and execution horizons are
held fixed for each evaluated configuration.

We post-train the 100k-step camera-action DROID checkpoint for 10,000 updates on four NVIDIA H800 GPUs (AdamW, batch size 128) with a prediction action chunk size $H=16$.

\section{EXPERIMENTS}
\label{sec:experiments}

\begin{figure}[t]
    \centering
    \includegraphics[width=1\columnwidth]{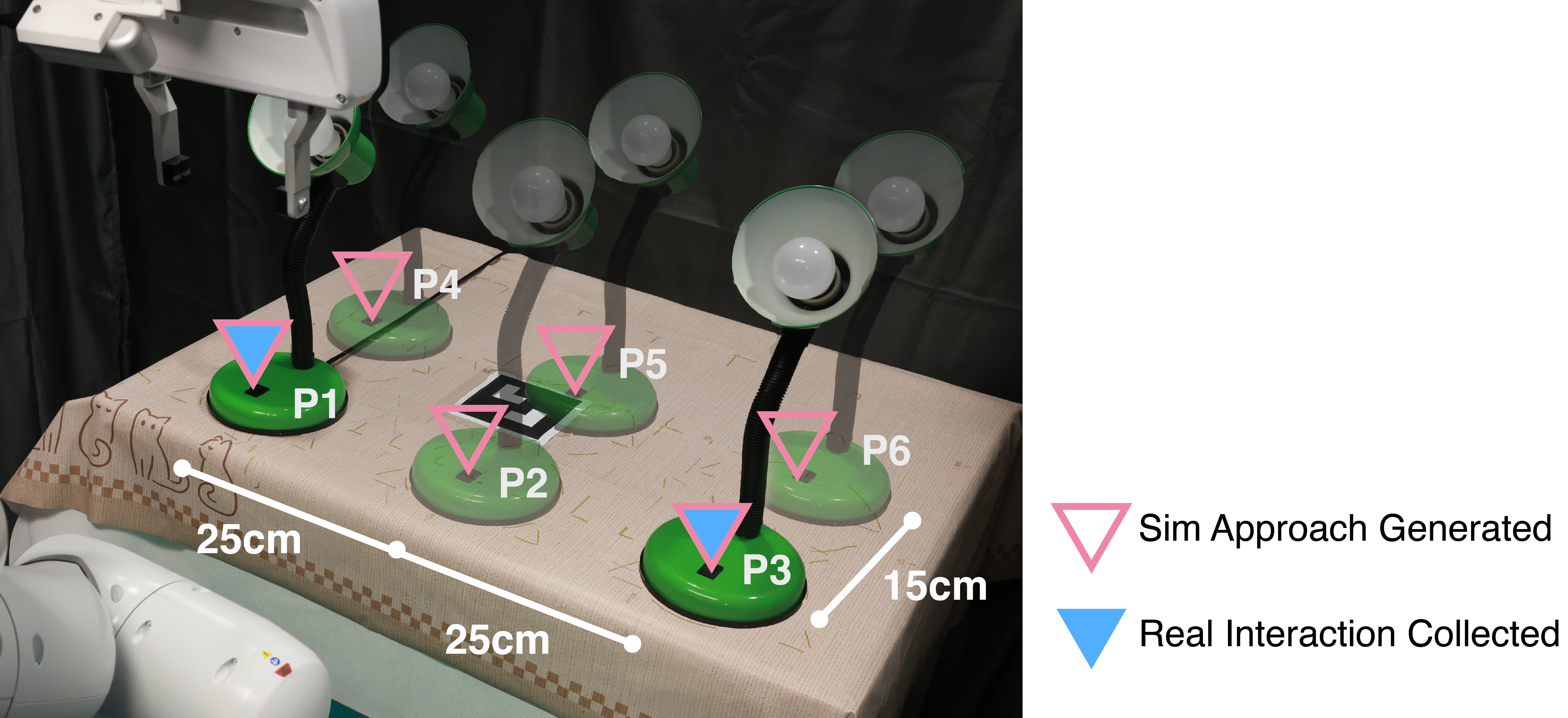}
    \caption{\textbf{Lamp-task placement protocol.} The six candidate lamp
    placements form a $2\times3$ grid with 25\,cm lateral and 15\,cm depth
    spacing. Simulated approach data cover all placements (pink markers), while
    real contact data are collected only at P1 and P3 (blue markers). The remaining
    four placements are unseen for real contact demonstrations and are used for evaluation.}
    \label{fig:exp-table-setup}
\end{figure}

We mainly evaluate \textbf{SARI} performance through the following research questions:

\textbf{Q1:} Can \textbf{SARI} improve spatial generalization to object placements without real interaction demonstrations while reducing real-data collection effort?

\textbf{Q2:} Does this phase-split recipe generalize across contact-rich tasks with different physical interaction dynamics?

\textbf{Q3:} Which appearance and action alignments are required for a single policy to stitch simulated approach data with real contact data?

\subsection{Experiment setup}
\label{sec:exp-setup}
We selected a suite of five tasks to test whether a policy can complete full interactions at unseen placements: pressing a standalone button, turning on a lamp, opening a drawer, opening a book, and pulling a tissue.
These tasks were chosen because they represent physical interactions that are difficult to model in a simulator.
Specifically, these involve spring resistance (buttons), static and kinetic friction (drawer), articulated lifting (book), and pinching deformable, compliant materials (tissue).

All five tasks share the same recipe: the robot must approach the object in free space and then make physical contact to execute the interaction.
Each task defines six nominal object placements on the workspace.
Real contact training data is restricted to only two of these placements, whereas simulated approaches cover all six.
We conduct all experiments using a Franka Research 3 with a Franka Hand, observed by two exterior cameras and a wrist camera in RealSense D435i.
A specialist policy is post-trained for each task.

The lamp is our primary evaluation testbed because its button is smaller than the standalone push button and its contact dynamics are difficult to reproduce in simulation.
Figure~\ref{fig:exp-table-setup} gives the lamp placements.
Real contact training is restricted to P1 and P3, while P2 and P4--P6 remain unseen for contact demonstrations.
Every evaluation starts from the same home configuration.
A lamp trial succeeds if the policy reaches the lamp and presses its button to turn it on within 40s, without intervention.

\begin{figure}[t]
    \centering
    \includegraphics[width=1\columnwidth]{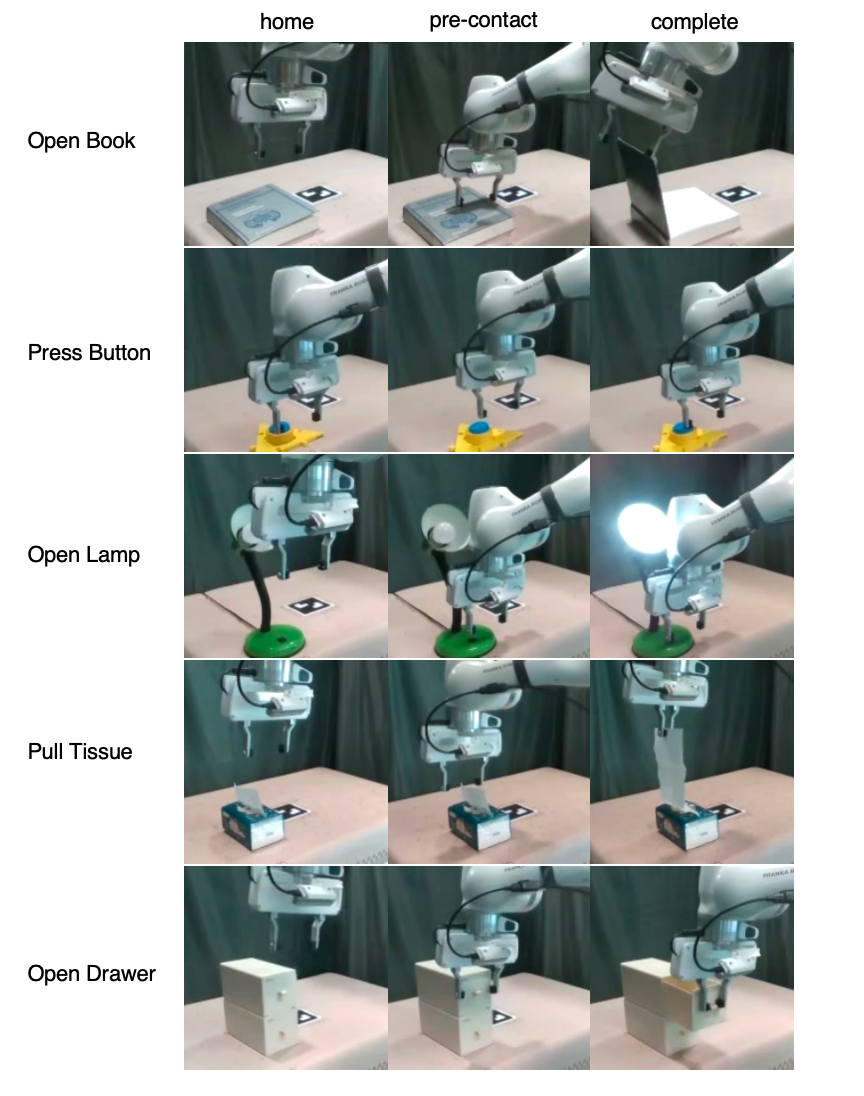}
    \caption{Representative real-world task trajectories. From top to
    bottom: press button, open lamp, open drawer, open book, and pull
    tissue. Each row progresses from the home configuration to task
    completion.}
    \label{fig:tasks}
\end{figure}

\subsection{Spatial Generalization and Real-Data Efficiency}
\label{sec:exp-main}
We compare our phase-split allocation against three baselines on the lamp task (Table~\ref{tab:main-results}). To construct the single-domain baselines, we collect full-task demonstrations progressing continuously from the home states to the light switch press. The Full Sim dataset is generated programmatically using ManiSkill, while the Full Real dataset is collected via human teleoperation. Unlike these full-task baselines, our phase-split method never observes a full-task demonstration, relying strictly on simulated approaches and real-world contact segments. Table~\ref{tab:episode-composition} summarizes the resulting data allocation.

To systematically evaluate task execution, we decompose evaluation into two sequential milestones: \textbf{Approach} (successfully navigating from the home configuration to the pre-contact region), and \textbf{Completion} (successfully toggling the lamp on). All four methods start from the same camera-action DROID-midtrained checkpoint and use an identical post-training compute budget.

\textbf{Phase-split allocation unlocks out-of-domain spatial generalization.}
As shown in Table~\ref{tab:main-results}, our phase-split method is the only approach involving real data that achieves nonzero full-task completion at every unseen placement. By explicitly separating the task phases, our recipe leverages simulated approaches for spatial generalization while reserving real data for contact execution.
Notably, P2 attains 40.0\% completion, higher than the other unseen placements. Although P2 is unseen with respect to real contact coverage, it lies spatially between the seen placements P1 and P3; we therefore conjecture that its relatively strong performance may reflect spatial interpolation rather than pure extrapolation.

\textbf{Full-task sim-real co-training fails to transfer simulated spatial generalization.}
Crucially, simply co-training the policy on a mixture of full-task simulation and full-task real data does not resolve the failure at unseen placements. Although the Full-Task Co-Train method approaches P2, it completes none of the unseen trials. This demonstrates a severe limitation of standard co-training: when full real-world demonstrations are available, the policy overfits to the narrow real-world spatial distribution and discards the broader generalization available in the simulated data.

\textbf{Pretrained vision-language-action models lack zero-shot spatial generalization.}
The Full Real baseline can approach both seen placements (P1, P3) in every trial, but complete only half of these trials and no unseen trial. Without explicit training data at unseen placements, the $\pi_{0.5}$ policy defaults to memorized trajectories, typically attempting to navigate to the nearest seen placement (e.g., driving toward P1 when the lamp is actually located at P4).

\begin{table}[t]
\small
\centering
\begingroup
\caption{\textbf{Five--task success rates (\%).} The lamp task is evaluated in ten trials per placement, while all other tasks are evaluated in four trials per placement.
Seen: P1/P3; Unseen: P2/P4--P6. Unseen placements have no real contact demonstrations.}
\label{tab:other-tasks}
\renewcommand{\arraystretch}{1.05}

\begin{tabularx}{\columnwidth}{
@{}
>{\raggedright\arraybackslash}X
>{\centering\arraybackslash}X
>{\centering\arraybackslash}X
@{}}
\toprule
Task & \shortstack{Seen} & \shortstack{Unseen} \\
\midrule
Book & 87.5 & 81.25 \\
Button & 87.5 & 75.0 \\
Tissue & 62.5 & 43.75 \\
Lamp & 50.0 & 27.5 \\
Drawer & 50.0 & 25.0 \\
\bottomrule
\end{tabularx}

\endgroup
\end{table}

\textbf{Simulation provides highly accurate geometric guidance.}
The Full Sim baseline can approach every seen and unseen placement in every trial. This shows that our digital twin accurately models the workspace geometry and that the policy can efficiently learn spatial positioning purely from simulated data. However, the idealized binary contact proxy does not yield real-world task completion, necessitating real-world data for reliable contact execution.

\subsection{Evaluation Across Contact-Rich Tasks}
\label{sec:exp-generalization}
\textbf{The phase-split recipe generalizes across varying physical interactions.}
To verify that our methodology is not limited to the lamp task, we apply the identical segmented collection recipe to book, standalone button, tissue, and drawer tasks. Representative task trajectories are shown in Figure~\ref{fig:tasks}. Real contact demonstrations cover P1/P3 for each task, while simulated approaches cover all six placements. As shown in Table~\ref{tab:other-tasks}, the success rate on unseen placements decreases from 81.25\% on book to 75.0\% on button, 43.75\% on tissue, 27.5\% on lamp, and 25.0\% on drawer. These results demonstrate that the recipe transfers across distinct contact dynamics without requiring a task-specific hierarchical controller.

\begin{figure}[t]
\centering
\includegraphics[width=\columnwidth]{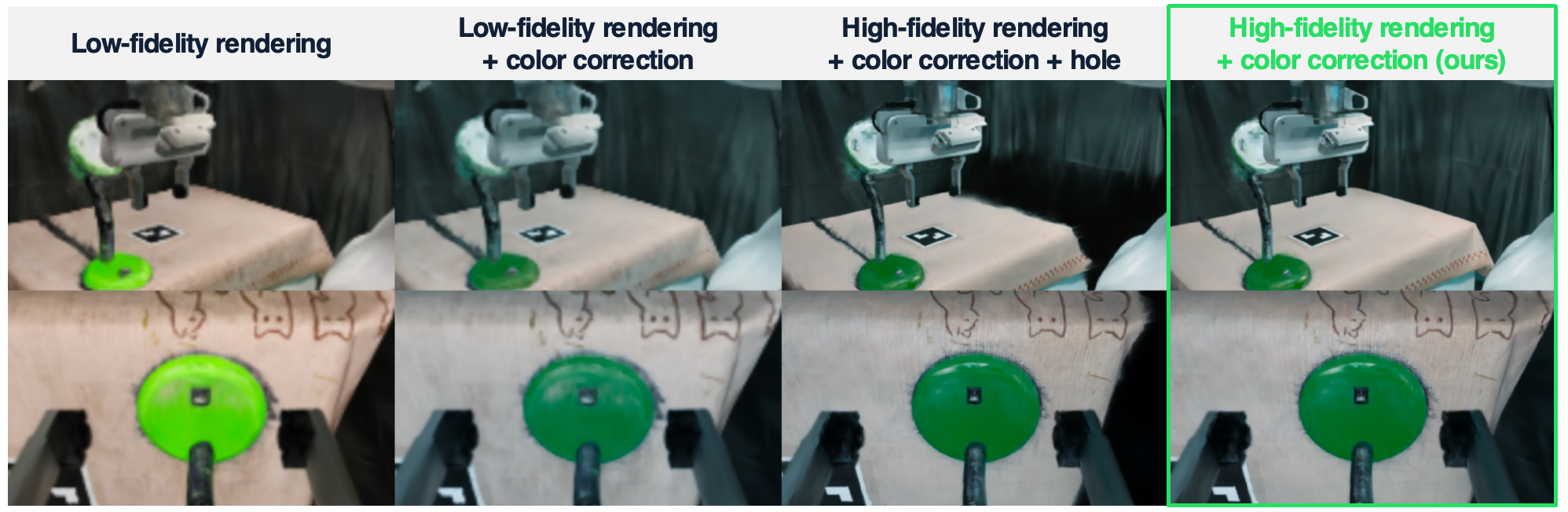}
\caption{Appearance configurations evaluated in Table~\ref{tab:appearance}.
From left to right: low-fidelity rendering; low-fidelity rendering with color
correction; high-fidelity rendering with color correction and corrupted GS; and
high-fidelity rendering with color correction and correct GS.}
\label{fig:appearance}
\end{figure}

\begin{table}[t]
\centering
\small
\begingroup
\caption{\textbf{Appearance-ablation success rates on the lamp task (\%).} We compare four visual
configurations under $\pi_{0.5}$ and GR00T N1.7 pretraining. Checkmarks denote
the listed condition. Each policy and configuration is evaluated in 60 trials (10 trials per placement across six object placements).}
\label{tab:appearance}
\renewcommand{\arraystretch}{1.05}

\begin{tabularx}{\columnwidth}{
@{}
>{\hsize=0.55\hsize\raggedright\arraybackslash}X
>{\hsize=0.45\hsize\centering\arraybackslash}X
>{\hsize=0.25\hsize\centering\arraybackslash}X
>{\hsize=0.35\hsize\centering\arraybackslash}X
>{\hsize=0.40\hsize\centering\arraybackslash}X
@{}}
\toprule
\multicolumn{3}{c}{Configuration} & \multicolumn{2}{c}{Success (\%)} \\
\cmidrule(lr){1-3} \cmidrule(lr){4-5}
Rendering & \shortstack{Color\\corr.} & \shortstack{GS\\corr.} & $\pi_{0.5}$~\cite{pi05} & \shortstack{GR00T\\N1.7~\cite{gr00tn1}} \\
\midrule
Low & -- & $\checkmark$ & 8.3 & 3.3 \\
Low & $\checkmark$ & $\checkmark$ & 13.3 & 11.7 \\
High & $\checkmark$ & -- & 18.3 & 13.3 \\
\mbox{\textbf{High (Ours)}} & $\checkmark$ & $\checkmark$ & \textbf{35.0} & \textbf{25.0} \\
\bottomrule
\end{tabularx}

\endgroup
\end{table}

\subsection{What Enables Sim-Real Phase Stitching?}
\subsubsection{Appearance Alignment}
\label{sec:exp-ablation}
Q3 examines the conditions required to stitch simulated approach data with real contacts. We first isolate three visual factors that determine whether simulated observations are sufficiently aligned with the real observation stream:

\begin{itemize}
\item \textbf{Rendering Fidelity}: Spherical-harmonics coefficients encode how a Gaussian's appearance changes with viewing angle. We compare a high-fidelity rendering that keeps degree-three spherical harmonics ($L=3$) at a high resolution against a low-fidelity baseline where harmonics are truncated to the constant, view-independent term ($L=0$).
\item \textbf{Photometric Color Correction}: The application of frozen, per-camera color maps and bilateral filtering, designed to minimize the sim-real color domain gap.
\item \textbf{Scene Completeness}: During development, attempts to clean the 3D environment by removing extraneous background Gaussians accidentally corrupted the scene, leaving holes in the render. We test robustness to this by comparing a \emph{corrupted-GS} condition against the \emph{correct-GS} baseline.
\end{itemize}

\textbf{Appearance-alignment result.} Bridging the visual domain gap requires targeted photometric corrections.
Figure~\ref{fig:appearance} visualizes the four configurations, which we compare across the same 240 lamp training episodes (as detailed in Table~\ref{tab:episode-composition}). As shown in Table~\ref{tab:appearance}, relying on low-fidelity rendering without color correction fails to bridge the domain gap. Applying photometric color correction is necessary to prevent the policy from keying off the domain rather than the task. Furthermore, missing background geometry explicitly hurts policy performance, underscoring the need for complete 3D scene reconstructions.

\subsubsection{Action Representation}
\label{sec:exp-action}
\textbf{Action-alignment result.} Camera-relative $\pi_{0.5}$ achieves the highest success rates among the evaluated configurations.
Beyond visual alignment, the policy must maintain action consistency when transitioning from simulated to real data. We compare four DROID-initialized policy configurations, all post-trained on our segmented task data: $\pi_{0.5}$ using joint velocity and joint position, GR00T N1.7 using end-effector-relative Cartesian actions, and our $\pi_{0.5}$ camera-relative actions (Table~\ref{tab:actionrep}).

\begin{table}[t]
\centering
\small
\caption{\textbf{Action-representation success rates on the lamp task (\%).}
EEF-relative uses GR00T N1.7~\cite{gr00tn1}; all other configurations use
$\pi_{0.5}$~\cite{pi05}, and camera-relative is ours.
Seen: P1/P3; Unseen: P2/P4--P6.}
\label{tab:actionrep}
\renewcommand{\arraystretch}{1.05}

\begin{tabularx}{\columnwidth}{
@{}
>{\raggedright\arraybackslash}X
>{\centering\arraybackslash}X
>{\centering\arraybackslash}X
@{}}
\toprule
Configuration & Seen & Unseen \\
\midrule
Joint velocity & 0.00 & 0.00 \\
Joint position & 0.00 & 0.00 \\
EEF-relative & 35.00 & 20.00 \\
Camera-relative & 50.00 & 27.50 \\
\bottomrule
\end{tabularx}

\end{table}

Both joint-space $\pi_{0.5}$ configurations fail on every trial. The GR00T N1.7 with EEF relative action achieves 35.0\% success on the seen placements and 20.0\% success on the unseen placements, while our camera-relative $\pi_{0.5}$ achieves 50.0\% success on the seen placements and 27.5\% success on the unseen placements. These results show that the camera-relative representation supports reliable sim-real phase stitching beyond joint space and EEF Cartesian action parameterizations.

Together, these appearance and action comparisons address Q3: the appearance
ablations assess rendering quality and photometric alignment, while the action
comparison identifies camera-relative $\pi_{0.5}$ as the best-performing
configuration among those evaluated.

\section{CONCLUSION}
\label{sec:conclusion}

We introduce \textbf{SARI} (Simulated Approach, Real Interaction), a
phase-split sim-and-real co-training framework for contact-rich manipulation.
Gaussian Splatting provides workspace-wide simulated approaches, while real
interactions are collected at a few placements. Photometric color alignment
and action-space standardization enable joint VLA post-training on these
trajectories. Experiments on five tasks show reduced real-data collection
and improved performance. On the lamp task, real demonstration time drops
from 28.03 to 18.42 minutes (34.3\%), with 11/40 (27.5\%) unseen task
completion versus 0/40 for all baselines. These results highlight the
importance of complementary phase allocation and sim-to-real alignment
for spatial generalization.

\textbf{Limitations.} \textbf{SARI} currently assumes fixed tabletop scenes,
a single-arm Franka robot, and two-stage approach--interaction tasks.
It requires manual scene reconstruction, camera calibration,
pre-contact-pose selection, and photometric reference selection.
Dynamic scenes, moving objects, multi-stage interactions, additional
robot embodiments, and automated preparation remain future work.

\section*{Acknowledgments}
ChatGPT and Gemini were used solely for language editing; the authors reviewed and take full responsibility for all content.

\bibliographystyle{IEEEtran}
\bibliography{IEEEabrv,references}

\end{document}